\documentclass{article}
\pdfoutput=1
\PassOptionsToPackage{numbers, compress}{natbib}
\usepackage[preprint]{neurips_2019}

\usepackage[utf8]{inputenc} % allow utf-8 input
\usepackage[T1]{fontenc}    % use 8-bit T1 fonts
\usepackage{hyperref}       % hyperlinks
\usepackage{url}            % simple URL typesetting
\usepackage{booktabs}       % professional-quality tables
\usepackage{amsfonts}       % blackboard math symbols
\usepackage{nicefrac}       % compact symbols for 1/2, etc.
\usepackage{microtype}      % microtypography
\usepackage{graphicx}
\usepackage{caption}
\usepackage{subfigure}
\usepackage{amsmath}

\title{Spectral-based Graph Convolutional Network for Directed Graphs}

\author{
  Yi Ma \\
  College of Intelligence and Computing \\
  Tianjin University \\
  \texttt{mayi@tju.edu.cn} \\
  \And
   Jianye Hao \\
   College of Intelligence and Computing \\
   Tianjin University \\
    \texttt{jianye.hao@tju.edu.cn} \\
    \And
   Yaodong Yang \\
   College of Intelligence and Computing \\
   Tianjin University \\
   \texttt{yydapple@gmail.com} \\
   \And
   Han Li\\
   Alibaba \\
   \texttt{lihan.lh@alibaba-inc.com} \\
   \And
   Junqi Jin\\
   Alibaba \\
   \texttt{junqi.jjq@alibaba-inc.com}\\
   \And
   Guangyong Chen \\
    Tencent\\
   \texttt{gycchen@tencent.com} \\
}

\begin{document}

\maketitle

\begin{abstract}
Graph convolutional networks(GCNs) have become the most popular approaches for graph data in these days because of their powerful ability to extract features from graph. GCNs approaches are divided into two categories, spectral-based and spatial-based. As the earliest convolutional networks for graph data, spectral-based GCNs have achieved impressive results in many graph related analytics tasks. However, spectral-based models cannot directly work on directed graphs. In this paper, we propose an improved spectral-based GCN for the directed graph by leveraging redefined Laplacians to improve its propagation model. Our approach can work directly on directed graph data in semi-supervised nodes classification tasks. Experiments on a number of directed graph datasets demonstrate that our approach outperforms the state-of-the-art methods. 

\end{abstract}

\section{Introduction}
In recent years, deep learning has achieved great success in many kinds of fields such as image classification, video processing and speech recognition. The data in these tasks is usually represented in the Euclidean space. However, there are many applications where data is generated from the non-Euclidean domain and is represented as graphs. This kind of data is known as graph data. A graph data structure consists of a finite set of vertices (also called nodes), together with a set of unordered pairs of these vertices for an undirected graph or a set of ordered pairs for a directed graph. These pairs are known as edges. Using the information of graph data, we can capture the interdependence among instances (nodes), such as citationship in papers network, friendship in social network and interactions in molecule network. For instance, in a papers citation network, papers are linked to each other via citationship and the papers can be classified into different areas. The graph data is very complex because of its irregularity. The complexity of graph data results that some important operations of deep learning are not applicable to non-Euclidean domain. For example, convolutional neural networks(CNNs) cannot use a convolution kernel of the same size to convolve graph data of such complex structure.

To handle the complexity of graph data, there have been many studies to design new models for graph data inspired by convolution networks, recurrent networks, and deep autoencoders. These models which incorporate neural architectures are known as graph neural networks. Graph neural networks are categorized into graph convolution networks, graph attention networks\cite{velivckovic2017graph} \cite{zhang2018gaan}, graph autoencoders\cite{Kipf2016Variational} \cite{wang2017mgae}, graph generative networks\cite{de2018molgan} \cite{li2018learning} and graph spatial-temporal networks\cite{li2017diffusion} according to Wu et al. \cite{wu2019comprehensive}. In these graph neural networks, graph convolution networks(GCNs) are the most important ones, which are the fundamental of other graph neural network models. One of the earliest work on GCNs is presented in Bruna et al. (2013), which develops a variant of graph convolution \cite{bruna2013spectral}. From then on, there have been many works to improve graph convolutional networks \cite{kipf2016semi} \cite{defferrard2016convolutional} \cite{henaff2015deep} \cite{li2018adaptive} \cite{levie2017cayleynets}. These GCNs approaches fall into two categories. One category of the GCNs approaches is spatial-based. These approaches directly perform the convolution in the graph domain by aggregating information of the neighbor nodes. The other category of the GCNs approaches is spectral-based. These approaches propose a variant of graph convolution methods based on spectral graph theory from the perspective of graph signal processing. Although spectral-based methods have more computational cost than spatial-based ones, they have more powerful ability to extract features from graph data.

As the earliest convolutional networks for graph data, spectral-based models have achieved impressive results in many graph related analytics tasks. However, spectral-based models are limited to work only on undirected graphs \cite{kipf2016semi}. So the only way to apply spectral-based models to directed graphs is to relax directed graphs to undirected ones, which would be unable to represent the actual structure of directed graphs. Some of the researchers combine the recurrent model and spectral-based GCN to process the temporal directed graphs \cite{pareja2019evolvegcn}, but they don't focus on the GCN's own structure. To the best of our knowledge, we are the first to make improvement of the spectral-based GCN layer's propagation model to make it adapted to directed graphs.

In this paper, we use a definition of the Laplacian matrix on directed graphs \cite{chung2005laplacians} to derive the propagation model's mathematical representation. We use feature decomposition and Chebyshev polynomials to approximate the representation of directed Laplacian matrix to get our propagation model. Then we use this propagation model to design our spectral-based GCNs for directed graphs. Our approach can work well on different directed graph datasets in semi-supervised nodes classification tasks and achieves better performance than the state-of-the-art spectral-based and spatial-based GCN methods. 

The remainder of this paper is organized as follows: Section \ref{back} introduces the theoretical motivation of the classic spectral-based GCNs; Section \ref{method} demonstrates the mathematical representation of Laplacians for directed graph and the models we construct in our methods; Section \ref{exp} demonstrates the details of our experiments on semi-supervised classification tasks; Concluding discussions and remarks are provided in Section \ref{dis} and Section \ref{conc}.

\section{Preliminaries}
\label{back}
Spectral-based GCNs are based on Laplacian matrix. For an undirected graph, suppose $A$ is the adjacency matrix of the graph, $D$ is a diagonal matrix of node degrees, $D_{ii}=\sum_{j}\left(A_{i,j}\right)$. A graph Laplacian matirx is defined as $L = D - A$. The normalized format of Laplacian matrix is defined as $L^{sym} =D^{-\frac{1}{2}} L D^{-\frac{1}{2}}=I_{n}-D^{-\frac{1}{2}} A D^{-\frac{1}{2}}$, which is a matrix representation of a graph in the graph theory and can be used to find many useful properties of a graph. $L^{sym}$ is symmetric and positive-semidefinite. With these properties, the normalized Laplacian matrix $L^{sym}$ can be factored as $L^{sym}=U{\Lambda}U^{T}$, where $U \in \mathbb{R}^{N \times N}$ is the matrix of eigenvectors ordered by eigenvalues and ${\Lambda}$ is the diagonal matrix of eigenvalues. 

\paragraph{Spectral Graph Convolutions} The spectral graph convolution operation is defined in the Fourier domain by computing the eigendecomposition of the graph Laplacian.  $x \in \mathbb{R}^{N}$  is the feature vector of graph's nodes. The graph Fourier transform to  $x$ is defined as $F(x)=U^{T} x$. The Fourier transform projects $x$ of input graph into the orthogonal space, which is equivalent to representing the arbitrary feature vector $x$ defined on the graph as a linear combination of the eigenvectors of the Laplacian matrix. The inverse graph Fourier transform is defined as $F^{-1}(\hat{x})=U \hat{x}$, where $\hat{x}$ is the output obtained by $x$ through the graph Fourier transform. Applying Convolution Theorem \cite{wiki:xxx} to the graph Fourier transform, the spectral convolutions on graphs are defined as the multiplication of $x$ with a filter $g \in \mathbb{R}^{N}$ in the Fourier domain:
\begin{equation}\label{eq:1}
x * g=F^{-1}(F(x) \odot F(g)) )=U\left(U^{T} x \odot U^{T} g\right) 
\end{equation}
where $*$ represents convolution operation and $\odot $ represents the Hadamard product. For two matrices $A$ and $B$ of the same dimension $m \times n$, the Hadamard product $(A \odot B)$ is a matrix of the same dimension as the operands, with elements given by$(A \odot B)_{i j}=(A)_{i j}(B)_{i j}$. By defining filter $g$ as $g_{\theta}= {diag} \left(U^{T} g\right)$, Equation \ref{eq:1} can be simplified as 
\begin{equation}\label{eq:2}
x * g_{\theta} = U g_{\theta} U^{T} x 
\end{equation}
Here we can understand $g_{\theta}$ as a function of the eigenvalues of $L^{sym}$, i.e. $g_{\theta}(\Lambda)$.

\paragraph{Chebyshev Spectral GCN} As we can see, multiplication with the eigenvector matrix $U$ from Equation \ref{eq:2} is computationally expensive. To solve this problem, Defferrard et al. \cite{defferrard2016convolutional} propose ChebNet  which uses Chebyshev polynomials of the diagonal matrix of eigenvalues to approximate $g_{\theta}$. ChebNet parametrizes $g_{\theta}$ to be a $K$ order polynomial of $\Lambda$:
\begin{equation}\label{eq:3}
g_{\theta}=\sum_{i=1}^{K} \theta_{i} T_{k}(\tilde{{\Lambda}})
\end{equation}
where $\tilde{\Lambda}=2 \Lambda / \lambda_{\max }-I_{\mathrm{N}}$ and $\lambda_{\max }$ denotes the largest eigenvalue of $L^{sym}$. The Chebyshev polynomials are defined recursively by $T_{k}(x)=2 x T_{k-1}(x)-T_{k-2}(x)$ with $T_{0}(x)=1$ and $T_{1}(x)=x$. Now the definition of a convolution of $x$ with a filter $g_{\theta}$  becomes:
\begin{equation}\label{eq:4}
x * g_{\theta} =\sum_{i=1}^{K} \theta_{i} T_{i}(\tilde{L}) x \end{equation}

where $\tilde{L}=2L^{sym} / \lambda_{\max }-I_{N}$. $\tilde{L}$ represents a rescaling of the graph Laplacian that maps the eigenvalues from $[0,\lambda_{\max }]$ to $[-1,1]$ since Chebyshev polynomial forms an orthogonal basis in $[-1,1]$.

\paragraph{First order of ChebNet(1stChebNet)} Kipf et al. \cite{kipf2016semi} propose a first-order approximation of ChebNet which assumes $K=1$ and $\lambda_{\max }=2$ to get a linear function. Equation \ref{eq:4} simplifies to: 
\begin{equation}\label{eq:5}
x * g_{\theta}=\theta_{0} x-\theta_{1} D^{-\frac{1}{2}} A D^{-\frac{1}{2}}{x} 
\end{equation}
And further assuming $\theta=\theta_{0}=-\theta_{1}$, the definition of graph convolution becomes
\begin{equation}\label{eq:6}
x * g_{\theta}=\theta\left(I_{N}+D^{-\frac{1}{2}} A D^{-\frac{1}{2}}\right) x 
\end{equation}
Because $I_{N}+D^{-\frac{1}{2}} A D^{-\frac{1}{2}}$ has eigenvalues in the range $[0,2]$, it may lead to exploding or vanishing gradients when used in a deep neural network model. To alleviate this problem, Kipf et al.  [12] use a renormalization trick $I_{N}+D^{-\frac{1}{2}} A D^{-\frac{1}{2}} \rightarrow \tilde{D}^{-\frac{1}{2}} \tilde{A} \tilde{D}^{-\frac{1}{2}}$, with $\tilde{A}=A+I_{N}$ and $\tilde{D}_{i i}=\sum_{j} \tilde{A}_{i j}$. It's a further simplification and it means adding a self-loop to each node in practice. Finally, we can generalize this definition of the graph convolution layer:
\begin{equation}\label{eq:7}
Z=\tilde{D}^{-\frac{1}{2}} \tilde{A} \tilde{D}^{-\frac{1}{2}} X \Theta 
\end{equation}
where $X \in \mathbb{R}^{N \times C}$ is with C-dimensional feature vector for every node, $\Theta \in \mathbb{R}^{C \times F}$ is a matrix of filter parameters and $Z \in \mathbb{R}^{N \times F}$ is the convolved result. The graph convolution defined by this format is localized in space and connects the spectral-based methods with spatial-based ones.

However, the above derivation is based on a premise that the Laplacian matrix is the representation for undirected graphs. It results that these spectral-based models are limited to work only on undirected graphs \cite{kipf2016semi}. The only way to handle directed edges is to relax directed graphs to undirected ones, which would be unable to represent the actual structure of directed graphs. To address this problem, we propose our spectral-based GCN method for directed graphs in the following section.

\section{Method}
\label{method}
Existing spectral-based GCNs methods cannot directly work on the directed graphs, but their powerful ability to extract features from graphs are impressive. It is expected that utilizing this ability of spectral-based GCNs in our work can improve the performance of our method. Besides, designing a spectral-based GCN is important for filling the gaps in the field of processing the directed graphs. Motivated by these, we design a spectral-based GCN method for the directed graph in our work.

In this section, we first give the definition of the Laplacians for directed graphs \cite{chung2005laplacians}, which is fundamental in spectral-based GCN. We then give the approximation of localized spectral filters on directed graphs using Chebyshev polynomials of the diagonal matrix of Laplacian's eigenvalues. Finally, we describe the models we use in our experiments.
\label{me}
\subsection{Laplacians for directed graphs}
Eigenvalues and eigenvectors are closely related to almost all major invariants of a graph, linking one extremal property to another. They play a central role in the fundamental understanding of graphs in spectral graph theory \cite{chung1997spectral}. The eigenvalues and eigenvectors of Laplacian matrix provide very useful information of graph. In a graph Laplacian, if two vertices are connected by an edge with a large weight, the values of the eigenvector at those locations are likely to be similar. The eigenvectors associated with larger eigenvalues oscillate more rapidly and are more likely to have dissimilar values on vertices connected by an edge with high weight. In addition, Laplacian matrix is a semi-positive symmetric matrix and the eigenvectors of the Laplacian matrix are a set of orthogonal basis in n-dimensional space, it's convenient to perform graph Fourier transform and inverse graph Fourier transform in practice as described in Section \ref{back}.
According to what we discussed above, the Laplacian matrix can represent the properties of graphs well and graph Laplacian eigenvectors can be used as filtering bases of GCN. In order to deduce the principal properties and structure of a graph from its graph spectrum, we choose to use Laplacian matrix for directed graphs to be the fundamental of our method. 

Suppose $G$ is a directed graph $G$ with vertex set $V (G)$ and edge set $E(G)$. For a directed edge $(u, v)$ in $E(G)$, we say that there is an edge $(u, v)$ from $u$ to $v$, or, $u$ has an out-neighbor $v$. The number of out-neighbors of $u$ is the out-degree of $u$, denoted by $d_{u}^{out}$. Using the same representation in Section \ref{back}, we can define $D_{\text {out}}= diag \left(\sum_{j}^{n} A_{i j}\right)$ as the out-degree matrix of a directed graph, where $A$ is the adjacency matrix(or weight matrix for weighted directed graph) of the directed graph. If there is a path in each direction between each pair of vertices of the graph $G$,  then this directed graph $G$ is called strongly connected.

\paragraph{Transition Probability Matrix} Assuming $P$ is a transition probability matrix, where $P(u, v)$  denotes the probability of moving from vertex $u$ to vertex $v$. For a given directed graph $G$ , a transition probability matrix $P$ is defined as
\begin{equation}\label{eq:8}
P(u, v)=\left\{\begin{array}{ll}{\frac{1}{d_{u}}} & {\text { if } (u, v)  \text { is an edge }} \\ {0} & {\text { otherwise. }}\end{array}\right.
\end{equation}
For a weighted directed graph with edge weights $w_{u v} \geq 0$, a transition probability matrix $P$ can be defined as being proportional to the corresponding weights and formally we have
\begin{equation}\label{eq:9}
P(u, v)=\left\{\begin{array}{ll}{\frac{w_{u v}}{\sum_{z} w_{u z}}} & {\text { if } (u, v)  \text { is an edge }} \\ {0} & {\text { otherwise. }}\end{array}\right.
\end{equation}
An unweighted directed graph is just a special case with weight having value 1 or 0. In practice, the transition probability matrix can be presented by 

\begin{equation}\label{eq:10}
P=D_{out}^{-1} A
\end{equation}
\paragraph{Perron Vector} The Perron-Frobenius Theorem \cite{horn2012matrix} states that an irreducible matrix with non-negative entries has a unique left eigenvector with all entries positive. This can be translated to language for directed graphs. Let $\rho$ denote the eigenvalue of the all positive eigenvector of the transition probability matrix $P$, $P$ of a strongly connected directed graph has a unique left eigenvector $\phi$ with $\phi(v)>0$ for all $v$ and 
\begin{equation}\label{eq:11}
\phi P=\rho \phi
\end{equation}
where $\phi$ is a row vector. According to the Perron-Frobenius Theorem, we have $\rho=1$ and all other eigenvalues of $P$ have absolute value at most 1. Then we normalize and choose $\phi_{norm}$ so that
\begin{equation}\label{eq:12}
\sum_{v} \phi_{norm}(v)=1
\end{equation}
We call $\phi_{norm}$ the Perron vector of $P$. For a strongly connected graph, $\phi_{norm}$ is a stationary distribution.
Define $\Phi=diag(\phi_{norm}(v))$. Using $\Phi$, we establish the Laplacians for directed graphs in the following paragraph.

\paragraph{Definition of Directed Laplacian} As described in Section \ref{back}, in undirected graphs, we have the definition of $L^{\text { sym }}$ and we can further derive this definition
\begin{equation}\label{eq:13}
L^{\text { sym }}=I-D^{-1 / 2} A D^{-1 / 2}=I-D^{1 / 2} P D^{-1 / 2}=I-\Phi^{1 / 2} P \Phi^{-1 / 2}
\end{equation}
Now we generalize this definition of undirected graphs to directed graph. We find the most important problem is that $P$ is not
symmetric in directed graph. So we use this following definition to guarantee that the normalized Laplacian is symmetric.
\begin{equation}\label{eq:14}
L^{sym}=I-\frac{1}{2}\left(\Phi^{1 / 2} P \Phi^{-1 / 2}+\Phi^{-1 / 2} P^{T} \Phi^{1 / 2}\right) 
\end{equation}

\subsection{Spectral GCN for Directed Graph}
\label{3.2}
As the Laplacian defined in Equation \ref{eq:14} is symmetric, we can calculate it's eigendecomposition as the filter. Then we approximate this filter using the Chebyshev polynomials and set it to first-order as we demonstrated in Section \ref{back}. Finally, we can derive the definition of the directed graph convolution layer:
\begin{equation}\label{eq:15}
Z= \frac{1}{2}\left(\tilde{\Phi}^{1 / 2} \tilde{P} \tilde{\Phi}^{-1 / 2}+\tilde{\Phi}^{-1 / 2} \tilde{P}^{T} \tilde{\Phi}^{1 / 2}\right)X \Theta 
\end{equation}
where adjacent matrix(weight matrix) $\tilde{A}$ used in this definition to derive $\tilde{P}$ and $\tilde{D}_{out}$ are added self-loop for each node. That is, $\tilde{A}=A+I_{N}$, $\tilde{D}_{out}=\sum_{j} \tilde{A}_{i j}$ , $\tilde{P}=\tilde{D}_{out}^{-1} \tilde{A}$ and $\tilde{\Phi}$ is calculated based on $\tilde{P}$. $X$ is feature vector for every node, $\Theta$ is a matrix of filter parameters, $Z$ is the convolved result.

\begin{figure}[htbp]
  \centering
  \caption{Details of DGCN propagation model.}
  \label{figure-1}
  \includegraphics[scale=0.38]{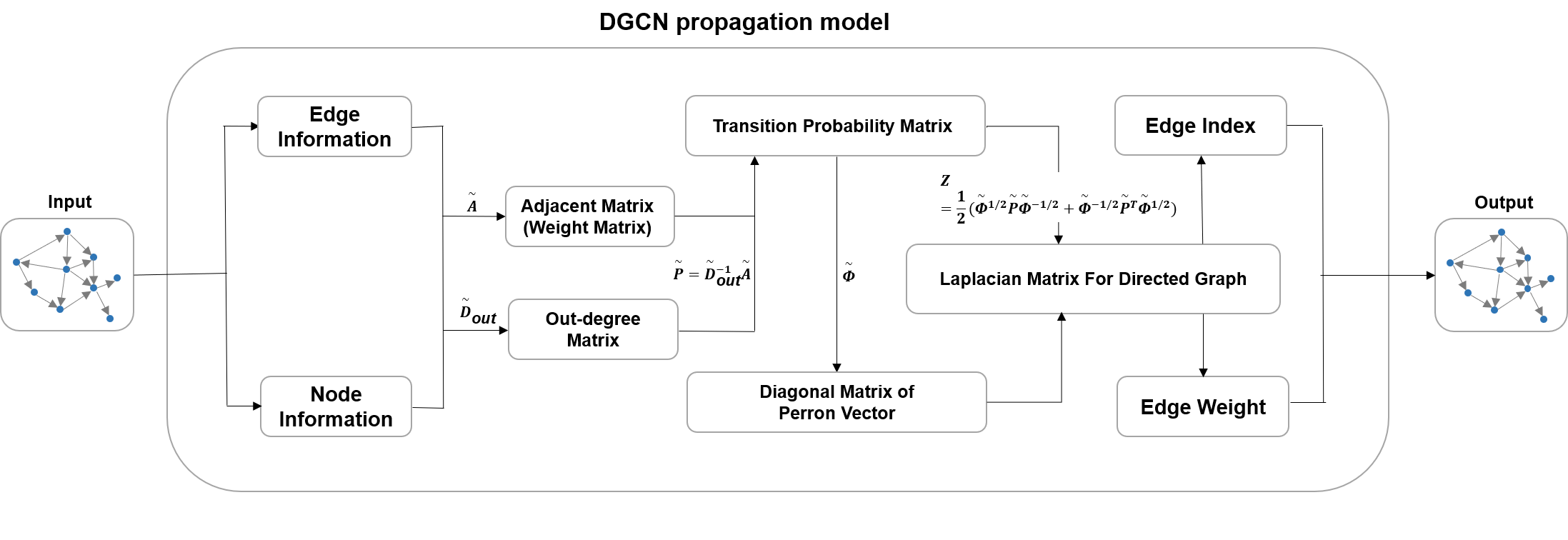}
\end{figure}

Now we get the propagation model for directed graph convolution of our method DGCN(Directed Graph Convolutional Network). The details of DGCN propagation model are shown in Figure \ref{figure-1}. The symbols in this figure represent the same meaning as defined in Equation \ref{eq:15}. Edge information and node information is obtained from the input. The edge index and edge weight represent the edge and its weight in the graph after processing in DGCN propagation model.  

\subsection{Models}
\label{3.3}
After introducing the propagation model, we design training models to solve the semi-supervised node classification for directed graph. In pre-processing step, we calculate $\hat{A}=\frac{1}{2}\left(\tilde{\Phi}^{1 / 2} \tilde{P} \tilde{\Phi}^{-1 / 2}+\tilde{\Phi}^{-1 / 2} \tilde{P}^{T} \tilde{\Phi}^{1 / 2}\right)$. Based on the conclusions in Section \ref{3.2}, we can naturally design models of multiple layers. Here we give a two-layer DGCN for example. 
\begin{equation}\label{eq:16}
Z=f(X, \hat{A})=softmax\left(\hat{A}ReLU\left(\hat{A} X W^{(0)}\right) W^{(1)}\right)
\end{equation}
where $X$ is the vectors of nodes' features. Note that $X$ doesn't contain information presented in $\hat{A}$, such as links between pages in a Wikipedia network. The neural network weights $W^{(0)}$ and $W^{(1)}$ are trained using gradient descent. In Equation \ref{eq:16}, $W^{(0)}$ is an input-to-hidden weight matrix and $W^{(1)}$ is a hidden-to-output weight matrix. The softmax activation function is $\operatorname{softmax}\left(x_{i}\right)=\frac{\exp \left(x_{i}\right)}{\sum_{i} \exp \left(x_{i}\right)} $. We evaluate the cross-entropy loss over all labeled examples:
\begin{equation}\label{eq:18}
\mathcal{L}=-\sum_{l \in \mathcal{Y}_{L}} \sum_{f=1}^{F} Y_{l f} \ln Z_{l f}
\end{equation}
where $Y_{i}$ denotes labels and $\mathcal{Y}_{L}$ is the set of node indices that have labels.
We also use dropout to reduce overfitting in our graph convolutional network.

\begin{figure}[htbp]
  \centering
  \caption{Architectures of our models.}
  \label{figure-2}
  \includegraphics[scale=0.53]{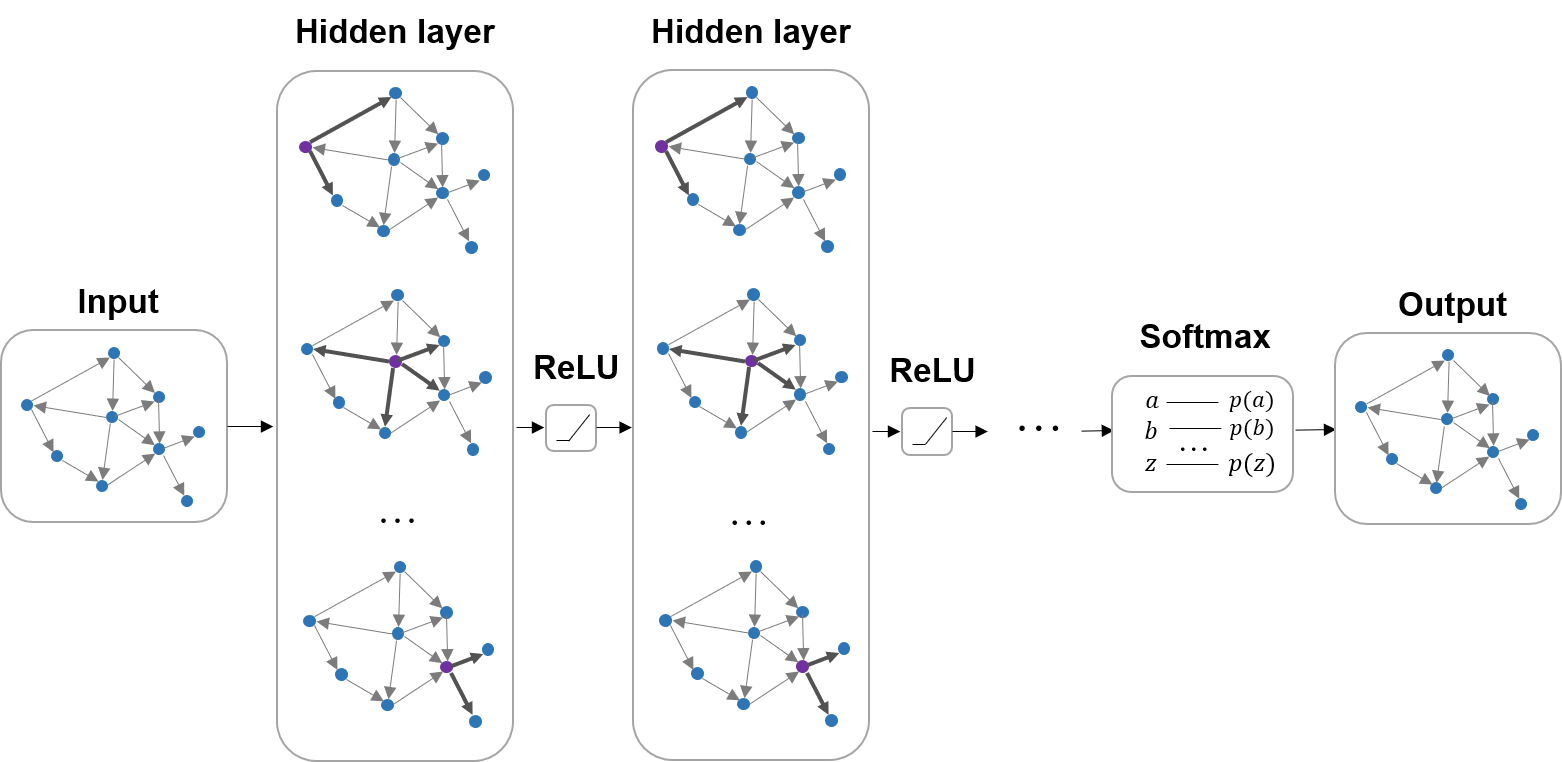}
\end{figure}

Considering the semi-supervised classification tasks of different difficulty level,  we design two models in our experiments. One is a two-layer model and the other is a three-layer model. The reason we use two-layer and three-layer model is to avoid overfitting along with the increasing number of parameters with deeper model depth as described in \cite{kipf2016semi}. Figure \ref{figure-2} shows the architectures of our models. Each hidden layer in the graph convolutional network is a DGCN propagation model.

\section{Experiments}
\label{exp}
We test our models in the semi-supervised nodes classification tasks on four different datasets. All the datasets in our experiments can be obtained from open sources. These datasets have different graph structures and belong to different kinds of networks(citation networks, hyperlink networks and email networks). It guarantees that the assessments based on these datasets are comprehensive and objective.

\subsection{Datasets}
Dataset statistics are summarized in Table ~\ref{table-1}. We introduce the number of total nodes and edges of each dataset. The nodes belong to different classes and we give the number of these classes. Nodes and edges of the largest strongly connected component(LSCC) are also showed in this table.  For all the datasets, we calculate the strongly connected component of the graphs and process the graphs into the edgelist format. The details of each dataset are given as follows.

\paragraph{Blogs} A directed network of hyperlinks among a large set of U.S. political weblogs from before the 2004 election \cite{Adamic:2005:PBU:1134271.1134277}. It includes blog political affiliation as metadata. Links between blogs were automatically extracted from a crawl of the front page of the blog. In addition, the authors drew on various sources (blog directories, and incoming and outgoing links and posts around the time of the 2004 presidential election) and classified 758 blogs as left-leaning and the remaining 732 as right-leaning.

\paragraph{Wikipedia} The hyperlink network of Wikipedia pages on editorial norms \cite{bradi16}, in 2015. Nodes are Wikipedia entries, and two entries are linked by a directed edge if one hyperlinks to the other. Editorial norms cover content creation, interactions between users, and formal administrative structure among users and admins. Metadata includes page information such as creation date, number of edits, page views and so on. The number of norm categories is also given.  

\paragraph{Email} The network was generated using email data from a large European research institution \cite{snapnets}. We have anonymized information about all incoming and outgoing email between members of the research institution. There is an edge $(u, v)$ in the network if person $u$ sent person $v$ at least one email. The emails only represent communication between institution members. The dataset also contains ground-truth community memberships of the nodes. Each individual belongs to exactly one of 42 departments at the research institute. 

\paragraph{Cora-cite} Citations among papers indexed by CORA, from 1998, an early computer science research paper search engine \cite{konect:2017:subelj_cora}. Nodes in CORA citation network represent scientific papers. If a paper $i$ cites a paper $j$ also in this dataset, then a directed edge connects $i$ to $j$. Papers not in the dataset are excluded. The papers are divided into 10 different computer science areas manually according to each paper's description. 

\begin{table}
  \caption{Datasets}
  \label{table-1}
  \centering
  \begin{tabular}{cccccc}
    \toprule
    Dataset  & Nodes   & Edges  & Nodes of LSCC  & Edges of LSCC  & Classes\\
    \midrule 
    Blogs     & 1490  & 19090  & 793  & 15783 & 2     \\
    Wikipedia & 1976  & 17235  & 1345 & 14601 & 10     \\
    Email     & 1005  & 25571  & 803  & 27429 & 42  \\
    Cora-cite & 23166 & 91500  & 3991 & 18007 & 10  \\
    \bottomrule
  \end{tabular}
\end{table}

\subsection{Set-up}
We follow the experimental setup in \cite{kipf2016semi}. In pre-processing, we calculate the largest strongly connected component of each dataset. For simple tasks(e.g. datasets with less than or equal to 10 classes), we design a two-layer model. For complicated tasks(e.g. Email dataset has more than 40 classes of nodes and less than 1000 nodes), we design a three-layer model to better extract graph data features. We use these two models for our four datasets. We train the models using about 10\%  of the nodes of the graph in each dataset following the settings of existing works \cite{kipf2016semi}. Then we use the rest of 90\% nodes as test datasets to evaluate prediction accuracy. For the node features, we concatenate a one-hot encoding of each node in the graph and the original features from the datasets.  In practice, we implement our method using PyTorch and PyTorch Geometric(A geometric deep learning extension library for PyTorch) \cite{fey2019fast}. The codes to reproduce our experiments will be published if our paper is accepted.

\subsection{Baselines}
 We compare with several state-of-the-art baselines methods, including spatial-based method\cite{morris2018weisfeiler}, spectral-based methods\cite{kipf2016semi}\cite{defferrard2016convolutional} and method combining with the attention mechanism\cite{velivckovic2017graph}. The first is the classic spectral-based 1stChebNet(GCN) \cite{kipf2016semi}. This is one of the best spectral-based GCN according to \cite{kipf2016semi}. The second is the Chebyshev spectral convolutional graph(ChebConv) \cite{defferrard2016convolutional}. The third method is the graph attention network(GAT) \cite{velivckovic2017graph}, which leverages masked self-attentional layers to address the shortcomings of classic GCN methods. The fourth is the graph neural network(GraphConv) \cite{morris2018weisfeiler}, which can take higher-order graph structures at multiple scales into account. In this method, we choose mean function to aggregate node features as described in their paper.

\subsection{Results}
Results of classification accuracy on test sets of our experiments are summarized in Table ~\ref{table-2}. We trained and tested our models on the datasets with different splitting of train sets and test sets. We report the mean accuracy and confidence interval of 20 runs with random weight initializations. For Blogs, Wikipedia and Cora-cite datasets, we use the two-layer model. For Email dataset, we use the three-layer model. For the same dataset, we use training model with the same architecture and parameters. The only difference is the propagation model of the convolution layer. 

As we can see in Table ~\ref{table-2}, our method outperforms the four baselines on four different datasets. The reason that our method achieves better performances may be described as followed. Our method makes use of the Laplacian designed for directed graphs, which has stronger ability to capture the connections between nodes of the network and to extract features from directed graphs.

The performances of all the methods are not so well on Cora-cite dataset and we believe there are three reasons. First, the Cora-cite dataset has 3991 nodes and only 18007 edges, it's a complex classification task. Second, the dataset has no node features, we have to construct a one-hot encoding of each node in the graph as the node features. Third, the classes of this dataset are manually divided into 10 areas according to each paper's description, it may cause some deviation from the ground truth.

\begin{table}
  \caption{Results of classification accuracy on test sets with 95\% confidence level(in percent)}
  \label{table-2}
  \centering
  \begin{tabular}{ccccc}
    \toprule
    Method  & Blogs   & Wikipedia  & Email  & Cora-cite  \\
    \midrule 
    GCN     & $89.38\pm 0.6$  & $62.98\pm 0.4$  & $54.06\pm0.4$  & $37.85\pm 0.3$  \\
   GraphConv   & $96.16\pm 0.6$  & $63.62\pm 0.3$  & $52.32\pm0.5$  & $37.98\pm 0.3$  \\
   GAT         & $91.26\pm0.7$  & $63.50\pm 0.4$  & $50.19\pm0.4$  & $38.28\pm 0.2$  \\
   ChebConv    & $88.92\pm 0.6$ & $61.71\pm 0.4$ & $44.36\pm0.5$  & $36.84\pm 0.3$ \\
    {\bf DGCN}(Ours)  & $\bf{97.09\pm 0.7}$  & $\bf{64.83\pm 0.4}$  & $\bf{57.63 \pm 0.4}$  & $\bf{38.78\pm 0.2}$  \\
    \bottomrule
  \end{tabular}
\end{table}

\section{Discussion}
\label{dis}
As demonstrated in the previous sections, our method for semi-supervised nodes classification of directed graphs outperforms several state-of-the-art methods. However, our method does have some limitations. 
First, the computational cost of our model increases with the graph size because our method needs to compute eigenvector of the transition probability matrix. It's a practical way to reduce the computational cost by implementing the matrix product using Coordinate Format(COO Format), but when paralleling or scaling to large graphs,  the computational cost of our spectral-based method is still a problem.
Second, our method also has to handle the whole graph at the same time, so the memory requirement is very high for spectral-based GCN method. The approximations of the large and densely connected graph can be very helpful as described in \cite{kipf2016semi}.
Third, our method is based on a premise that the input directed graph of our DGCN model should be strongly connected. According to this, we should calculate the largest strongly connected components of each dataset, which can cause some nodes to be removed from the original graph.

\section{Conclusion and Future Work}
\label{conc}
In this paper, we propose a novel method to design the propagation model of spectral-based GCN layer to adapt to directed graphs. Experiments on a number of directed network datasets suggest that our method can work directly on the directed graph in the semi-supervised nodes classification tasks. Our method outperforms several state-of-the-art baseline methods, including spatial-based methods, spectral-based methods and methods combining with the attention mechanism. 

In the future, there are several potential improvements and extensions to our work. For example, overcoming the practical problems described in Section \ref{dis} to reduce the computing cost and to handle graph in batch sizes can be a challenge in future work. We also believe it's feasible to combine other techniques like attention mechanism with our method to improve the performances on more datasets. In addition, combining GCN for directed graphs and reinforcement learning in multi-agent systems may be an attractive idea.

\small

\bibliographystyle{plain}
\bibliography{bibliography.bib}

\begin{thebibliography}{10}

\bibitem{konect:2017:subelj_cora}
Cora citation network dataset -- {KONECT}, April 2017.

\bibitem{Adamic:2005:PBU:1134271.1134277}
Lada~A. Adamic and Natalie Glance.
\newblock The political blogosphere and the 2004 u.s. election: Divided they
  blog.
\newblock In {\em Proceedings of the 3rd International Workshop on Link
  Discovery}, LinkKDD '05, pages 36--43, New York, NY, USA, 2005. ACM.

\bibitem{bruna2013spectral}
Joan Bruna, Wojciech Zaremba, Arthur Szlam, and Yann LeCun.
\newblock Spectral networks and locally connected networks on graphs.
\newblock In {\em International Conference on Learning Representations (ICLR)},
  2014.

\bibitem{chung2005laplacians}
Fan Chung.
\newblock Laplacians and the cheeger inequality for directed graphs.
\newblock {\em Annals of Combinatorics}, 9(1):1--19, 2005.

\bibitem{chung1997spectral}
Fan~RK Chung and Fan~Chung Graham.
\newblock {\em Spectral graph theory}.
\newblock Number~92. American Mathematical Soc., 1997.

\bibitem{de2018molgan}
Nicola De~Cao and Thomas Kipf.
\newblock Molgan: An implicit generative model for small molecular graphs.
\newblock {\em arXiv preprint arXiv:1805.11973}, 2018.

\bibitem{defferrard2016convolutional}
Micha{\"e}l Defferrard, Xavier Bresson, and Pierre Vandergheynst.
\newblock Convolutional neural networks on graphs with fast localized spectral
  filtering.
\newblock In {\em Advances in neural information processing systems}, pages
  3844--3852, 2016.

\bibitem{fey2019fast}
Matthias Fey and Jan~Eric Lenssen.
\newblock Fast graph representation learning with pytorch geometric.
\newblock {\em arXiv preprint arXiv:1903.02428}, 2019.

\bibitem{bradi16}
Bradi Heaberlin and Simon DeDeo.
\newblock The evolution of {W}ikipedia's norm network.
\newblock {\em Future Internet}, 8(2):14, 2016.

\bibitem{henaff2015deep}
Mikael Henaff, Joan Bruna, and Yann LeCun.
\newblock Deep convolutional networks on graph-structured data.
\newblock {\em arXiv preprint arXiv:1506.05163}, 2015.

\bibitem{horn2012matrix}
Roger~A Horn and Charles~R Johnson.
\newblock {\em Matrix analysis}.
\newblock Cambridge university press, 2012.

\bibitem{kipf2016semi}
T.~N. Kipf and M.~Welling.
\newblock Semi-supervised classification with graph convolutional networks.
\newblock In {\em Proceedings of the International Conference on Learning
  Representations}, 2017.

\bibitem{Kipf2016Variational}
Thomas~N Kipf and Max Welling.
\newblock Variational graph auto-encoders.
\newblock {\em arXiv preprint arXiv:1611.07308}, 2016.

\bibitem{snapnets}
Jure Leskovec and Andrej Krevl.
\newblock {SNAP Datasets}: {Stanford} large network dataset collection.
\newblock \url{http://snap.stanford.edu/data}, June 2014.

\bibitem{levie2017cayleynets}
Ron Levie, Federico Monti, Xavier Bresson, and Michael~M Bronstein.
\newblock Cayleynets: Graph convolutional neural networks with complex rational
  spectral filters.
\newblock {\em IEEE Transactions on Signal Processing}, 67(1):97--109, 2017.

\bibitem{li2018adaptive}
Ruoyu Li, Sheng Wang, Feiyun Zhu, and Junzhou Huang.
\newblock Adaptive graph convolutional neural networks.
\newblock In {\em Thirty-Second AAAI Conference on Artificial Intelligence},
  2018.

\bibitem{morris2018weisfeiler}
Christopher Morris, Martin Ritzert, Matthias Fey, William~L Hamilton, Jan~Eric
  Lenssen, Gaurav Rattan, and Martin Grohe.
\newblock Weisfeiler and leman go neural: Higher-order graph neural networks.
\newblock {\em arXiv preprint arXiv:1810.02244}, 2018.

\bibitem{pareja2019evolvegcn}
Aldo Pareja, Giacomo Domeniconi, Jie Chen, Tengfei Ma, Toyotaro Suzumura,
  Hiroki Kanezashi, Tim Kaler, and Charles~E Leisersen.
\newblock Evolvegcn: Evolving graph convolutional networks for dynamic graphs.
\newblock {\em arXiv preprint arXiv:1902.10191}, 2019.

\bibitem{velivckovic2017graph}
Petar Veli{\v{c}}kovi{\'c}, Guillem Cucurull, Arantxa Casanova, Adriana Romero,
  Pietro Lio, and Yoshua Bengio.
\newblock Graph attention networks.
\newblock In {\em Proceedings of the International Conference on Learning
  Representations}, 2017.

\bibitem{wang2017mgae}
Chun Wang, Shirui Pan, Guodong Long, Xingquan Zhu, and Jing Jiang.
\newblock Mgae: Marginalized graph autoencoder for graph clustering.
\newblock In {\em Proceedings of the 2017 ACM on Conference on Information and
  Knowledge Management}, pages 889--898. ACM, 2017.

\bibitem{wiki:xxx}
{Wikipedia contributors}.
\newblock Convolution theorem --- {Wikipedia}{,} the free encyclopedia, 2019.
\newblock [Online; accessed 14-May-2019].

\bibitem{wu2019comprehensive}
Zonghan Wu, Shirui Pan, Fengwen Chen, Guodong Long, Chengqi Zhang, and Philip~S
  Yu.
\newblock A comprehensive survey on graph neural networks.
\newblock {\em arXiv preprint arXiv:1901.00596}, 2019.

\bibitem{li2018learning}
C.~Dyer R.~Pascanu Y.~Li, O.~Vinyals and P.~Battaglia.
\newblock Learning deep generative models of graphs.
\newblock In {\em Proceedings of the International Conference on Machine
  Learning}, 2018.

\bibitem{li2017diffusion}
C.~Shahabi Y.~Li, R.~Yu and Y.~Liu.
\newblock Diffusion convolutional recurrent neural network: Data-driven traffic
  forecasting.
\newblock In {\em Proceedings of International Conference on Learning
  Representations}, 2018.

\bibitem{zhang2018gaan}
Jiani Zhang, Xingjian Shi, Junyuan Xie, Hao Ma, Irwin King, and Dit-Yan Yeung.
\newblock Gaan: Gated attention networks for learning on large and
  spatiotemporal graphs.
\newblock In {\em Proceedings of the Uncertainty in Artificial Intelligence},
  2018.

\end{thebibliography}
\end{document}